\documentclass[11pt,a4paper]{article}
\usepackage[hyperref]{emnlp2020}
\usepackage{times}
\usepackage{latexsym}

\usepackage{url}

\aclfinalcopy % Uncomment this line for the final submission
\def\aclpaperid{62} %  Enter the acl Paper ID here

\usepackage[utf8]{inputenc}
\usepackage{amssymb}
\usepackage{amsmath}
\usepackage{algorithm}

\usepackage{bm}
\usepackage{multirow}
\usepackage{tablefootnote}

\usepackage{float}
\usepackage{subcaption}
\usepackage{mathtools, nccmath}
\usepackage{times}
\usepackage{latexsym}
\usepackage{array}
\usepackage{url}
\usepackage{emnlp2020}

\title{Addressing Exposure Bias With Document Minimum Risk Training: Cambridge at the WMT20 Biomedical Translation Task}
  
\author{Danielle Saunders \and Bill Byrne\\
\\
    Department of Engineering, University of Cambridge, UK  \\}

\begin{document}

\allowdisplaybreaks
\maketitle
\begin{abstract}
The 2020 WMT Biomedical translation task evaluated Medline abstract translations. This is a small-domain translation task, meaning limited relevant training data with very distinct style and vocabulary. Models trained on such data are susceptible to exposure bias effects, particularly when training sentence pairs are imperfect translations of each other. This can result in poor behaviour during inference if the model learns to neglect the source sentence.  

The UNICAM entry addresses this problem  during fine-tuning using a robust variant on Minimum Risk Training. We contrast this approach with  data-filtering  to remove `problem' training examples. Under MRT fine-tuning we obtain good results for both directions of English-German and  English-Spanish biomedical translation. In particular we achieve the best English-to-Spanish translation result and second-best Spanish-to-English result,  despite using only single models with no ensembling.
\end{abstract}

\section{Introduction}
Neural Machine Translation (NMT) in the biomedical domain presents challenges in addition to general domain translation. Text often contains specialist vocabulary and follows specific stylistic conventions. For this task fine-tuning generic pre-trained models on smaller amounts of biomedical-specific data can lead to strong performance, as we found in our 2019 biomedical submission \citep{saunders-etal-2019-ucam}. For our WMT 2020 submission we start with strong single models from that 2019 submission and fine-tune them exclusively on the small Medline abstracts training sets  \citep{bawden-etal-2019-findings}. This allows fast training on very relevant training data, since the test set is also made up of Medline abstracts. 

However, fine-tuning on relevant but small corpora has pitfalls. The small number of training examples exacerbates the effect of any noisy or poorly aligned sentence pairs. We treat this as a form of exposure bias, in that model overconfidence in training data results in poor translation hypotheses at test time.

Our contributions in this system paper are:
\begin{itemize}
    \item A discussion of exposure bias in the form of imperfect training data, focusing on the biomedical domain.
    \item An exploration of straightforward ways to mitigate exposure bias via data preparation and training objective.
    \item A discussion of our 2020 Biomedical task results for single models fine-tuned on small, domain-specific  data sets.
\end{itemize}

\subsection{Exposure bias in the biomedical domain}

\begin{table*}[ht]
\centering
\small
\begin{tabular}{|p{2.4cm}|p{12.6cm}|}\hline
English source & [Associations of work-related strain with subjective sleep quality and individual daytime sleepiness].\\
Human translation & [Zusammenhang von arbeitsbezogenen psychischen Beanspruchungsfolgen mit subjektiver Schlafqualität und individueller Tagesschläfrigkeit.] \\

\hline

MLE  & Zusammenfassung. \\
%MLE, removed square brackets &Assoziationen arbeitsbedingter Belastungen mit subjektiver Schlafqualität und individueller Tagesschläfrigkeit.\\
MRT  & [Assoziationen arbeitsbedingter Belastung mit subjektiver Schlafqualität und individueller Tagesschläfrigkeit]. \\
%MRT, removed square brackets &  Assoziationen arbeitsbedingter Belastungen mit subjektiver Schlafqualität und individueller Tagesschläfrigkeit.\\
\hline
English source & [Effectiveness of Upper Body Compression Garments Under Competitive Conditions: A Randomised Crossover Study with Elite Canoeists with an Additional Case Study]. \\
Human translation & [Effektivität von Oberkörperkompressionsbekleidung unter Wettkampfbedingungen: eine randomisierte Crossover-Studie an Elite-Kanusportlern mit einer zusätzlichen Einzelfallanalyse.]\\
\hline
MLE  & Eine randomisierte Crossover-Studie mit Elite-Kanuten mit einer Additional Case Study wurde durchgeführt. \\
%MLE, removed square brackets & Effektivität von Oberkörper-Kompressionsbekleidungsstücken unter kompetitiven Bedingungen: Eine randomisierte Crossover-Studie mit Spitzenkanuten mit einer zusätzlichen Fallstudie. \\
MRT  & Eine randomisierte Crossover-Studie mit Elite-Kanüsten mit einer Additional Case Study hat zur Wirksamkeit von Oberkörperkompressionsbekleidung unter kompetitiven Bedingungen geführt. \\

\hline
\end{tabular}
\caption{Two sentence from the English-German 2020 test set with hypothesis translations from various models, demonstrating the effects of exposure bias from training on imperfectly aligned training sentences. The first MLE example output is completely unrelated to the source sentence, but the second MLE translation is more misleading.}\label{tab:exposurebias}
\end{table*}
Exposure bias for an autoregressive sequence decoder  refers to a discrepancy between decoder conditioning during training and inference \citep{bengio2015scheduled, ranzanto16sequencelevel}. During training the decoder generates a hypothesis for the $t^{th}$ output token $\hat{y_t}$ conditioned on $y_{1:t-1}$, the gold target sequence prefix.  During inference, the gold target $y$ is unavailable, and $\hat{y_t}$ is conditioned instead on the hypothesis prefix $\hat{y}_{1:t-1}$.

Previous work has interpreted the risk of exposure bias primarily in terms of the model over-relying on correct gold target translations, resulting in error propagation when mistakes are made during inference. We take a different view, focusing on mistakes in the training data which harm the model through teacher-forcing exposure and cause it to make related mistakes during inference.

We identify a specific feature of the Medline abstract training data which caused noticeable translation errors. The data contains instances in which either the source or target sentence contains the correct translation of the other sentence, but adds information  that is not found in translation. For example, the following sentence appears in the English side of en-de Medline abstract training data:

\emph{[The effects of Omega-3 fatty acids in clinical medicine]. Effects of Omega-3 fatty acids (n-3 FA) in particular on the development of cardiovascular disease (CVD) are of major interest.}

Its corresponding German sentence is 

\emph{Der Nutzen von Omega-3-Fettsäuren (n-3-FS) in der Medizin, hauptsächlich in der Prävention kardio- und zerebrovaskulärer Erkrankungen, wird aktuell intensiv diskutiert.} (Translated: `The uses of Omega-3 fatty acids in medicine, especially in prevention of cardiovascular and cerebrovascular diseases, are currently heavily discussed.')

Some of the English sentence is present in the German translation, but the square-bracketed article title is not. In this example it might be possible to remove only the segment in square brackets, but in other examples there is even less overlap, while source and target sentences may still be related and therefore challenging to filter. For example, the following English and German sentences also correspond with still less overlap:

\emph{[Conflict of interest with industry--a survey of nurses in the field of wound care in Germany , Australia and Switzerland]. Background.}

\emph{Hintergrund: Pflegende werden zunehmend von der Industrie umworben.} (Translated: `Background: Nurses are being increasingly courted by industry.')

These examples are quite frequent in Medline abstract data, especially in the form of titles. It is common to insert the English title of a non-English article into its translation, marked with square brackets \citep{patrias2007citing}. The marked title is not present in the original article. Consequently models trained on English source sentences with titles can behave erratically when given sentences with square-bracketed titles at test time: an exposure bias effect.

One possible approach to this problem is aggressively filtering sentences which may be poorly aligned. However, with such a small training set, this risks losing valuable examples of domain-specific source and target language. We hypothesise that such filtering is not the only way to reduce the effects during inference. Instead, we propose an approach in terms of the parameter fine-tuning scheme with  Minimum Risk Training (MRT). \citet{wang-sennrich-2020-exposure} have recently shown MRT as effective for combating exposure bias in the context of domain shift --  test sentences which are very different from the training data. We propose that MRT is also more robust against exposure to misaligned training data.

The examples in Table \ref{tab:exposurebias} show the different behaviour of MLE and MRT in such cases.  In the first example, the MLE hypothesis is unrelated to the source sentence, while the MRT output is relevant. In the second example, the MLE output is more plausible and therefore misleading, as it still misses the first clause which the MRT hypothesis covers. Both MLE and MRT hypotheses are phrased like opening sentences rather than titles, and both feature the untranslated phrase `Additional Case Study': while MRT may be more robust, it is not immune to exposure bias.

We note that title translations may not exist in the human reference. In these cases failure to translate the title will not negatively impact BLEU. However, we argue a biomedical translation model should be able to translate such sentences if required. It is also important to note that title translations are not the only case of inexact training pairs, but are simply easily identifiable. 

\subsection{Document MRT}
\begin{figure}[ht]
\centering
\small
\includegraphics[width=\linewidth]{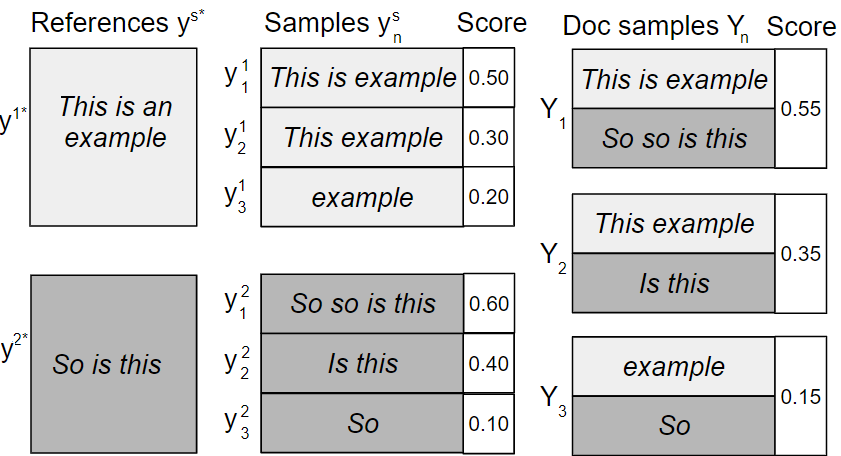}
\caption{Two MRT schemes with an $S=2$ sentence minibatch and $N=3$ samples / sentence. In standard MRT (middle) each sample has a score, e.g. sBLEU. For doc-MRT (right) samples are sorted into minibatch-level `documents', each with a combined score, e.g. document BLEU. Doc-MRT scores are less sensitive to individual samples, increasing robustness.}
\label{fig:mrt}
\end{figure}

Minimum Risk Training (MRT)  aims to minimize the expected cost  between $N$ sampled target sequences $\bm{y_n^{(s)}}$ and the corresponding gold reference sequence $\bm{y}^{(s)\ast}$ for the $S$ sentence pairs in each minibatch. For translation MRT is usually applied using a sentence-level BLEU (sBLEU) score corresponding to cost function $1 - \text{sBLEU}$, and sentence samples are generated by auto-regressive sampling with temperature $\tau$ during training  \citep{shen2016minimum}. Hyperparameter $\alpha$ controls sharpness of the distribution over samples. While MRT permits  training from scratch, in practice it is exclusively used to fine-tune models. 

Doc-MRT is a recently proposed MRT variant which changes sentence cost function to a document cost function, $D(.)$ \citep{saunders-etal-2020-using}. $D$ measures costs between minibatch-level `documents' $Y^*$ and $Y_n$.  $Y^*$ is formed of all $S$ reference sentences in the minibatch, and  $Y_n$ is one of $N$ sample `documents' each formed of one sample from each sentence pair $( \bm{x}^{(s)}, \bm{y}^{(s)\ast})$.  This permits MRT under document-level scores like BLEU, instead of sBLEU. The $n^{th}$ sample for the $s^{th}$ sentence in the minibatch-level document, $\bm{y}_n^{(s)}$, contributes the following term to the overall gradient:
$$
\frac{\alpha}{N} \sum_{Y: \bm{y}^{(s)} = \bm{y}^{(s)}_n} D(Y, Y^*)   \nabla_\theta  \log P(\bm{y}^{(s)}_n|\bm{x}^{(s)};\theta)  
%\frac{1}{N}\hspace*{-3ex} \sum_{Y: y^{(s)} = y^{(s)}_n} D(Y, Y^*)  
$$
In other words the gradient of each sample is weighted by the aggregated document-level scores for documents in which the sample appears.

Figure \ref{fig:mrt} gives a toy example of doc-MRT scoring samples in context. Document-level metrics aggregate scores across sentence samples, meaning a minibatch with some good samples and some poor samples will not have extreme score variation. Doc-MRT is therefore less sensitive than standard MRT to variation in individual samples. 

Doc-MRT has been shown to give better performance than standard MRT for small datasets with a risk of over-fitting, as well as improved robustness to small $N$. More discussion of these results and a derivation of the document-level loss function can be found in \citet{saunders-etal-2020-using}. Since we are attempting fine-tuning on small datasets and since $N$ is a limiting factor for MRT on memory-intensive large models, the biomedical task is an appropriate application for doc-MRT.

\begin{table*}[ht]
\centering
\small
\begin{tabular}{|l|l|l|l|l|l|}\hline
& \textbf{Phase}  & \textbf{Datasets} & \textbf{Sentence pairs} & \textbf{Dev datasets} & \textbf{Sentence pairs}\\ \hline
  \multirow{9}{*}{en-es}& \multirow{5}{*}{Pre-training}& UFAL Medical\footnotemark &639K & \multirow{5}{*}{Khresmoi\footnotemark} & \multirow{5}{*}{1.5K} \\
 & &Scielo\footnotemark & 713K &&  \\
 & & Medline titles\footnotemark& 288K&&  \\ 
 & & Medline abstracts & 83K && \\
 & & Total  &1723K / \textbf{1291K}&& \\
  \cline{2-6}
& Fine-tuning & Medline abstracts & 83K / \textbf{67.5K} & Biomedical19 & 800\\

 \hline
 \multirow{4}{*}{en-de} &  \multirow{3}{*}{Pre-training} & UFAL Medical  &2958K& Khresmoi&  1.5K\\
 && Medline abstracts & 33K &Cochrane\footnotemark &467 \\
  & & Total  &2991K / \textbf{2156K}& & \\
  \cline{2-6}
& Fine-tuning & Medline abstracts & 33K / \textbf{28.6K} & Biomedical19 & 800\\
\hline
\end{tabular}
\caption{Biomedical training and validation data used in the evaluation task. For both language pairs identical data was used in both directions. Bolded numbers are totals after filtering}\label{tab:data}
\end{table*}

\subsection{Related work}
Fine-tuning general models on domain-specific datasets has become common in NMT. Simple transfer learning on new data can adapt a general model to in-domain data \citep{luong2015stanford}. Mixed fine-tuning where some original data is combined with the new data avoids reduced performance on the original data-set \citep{chu2017empirical}. We are only interested in performance on one domain, so use simple transfer learning.

For this task, we specifically fine-tune on a relatively small dataset. Adaptation to very small, carefully-chosen domains has been explored for speaker-personalized translation  \citep{michel-neubig-2018-extreme} , and to reduce gender bias effects \citep{saunders-byrne-2020-reducing} while maintaining general domain performance. We wish to adapt to a very specific domain without need to maintain good general domain performance, but must avoid overfitting. Related approaches include fine-tuning a separate model for each test sentence \citep{li-etal-2018-one, farajian-etal-2017-multi} or test document \citep{xu2019lexical, kothur-etal-2018-document}. We choose to train a single model for all test sentences in a language pair, but improve the robustness of that model to overfitting and exposure bias using MRT. 

MRT has been widely applied to NMT in recent years \citep{shen2016minimum, neubig-2016-lexicons, edunov2018classical}. In particular, \citet{wang-sennrich-2020-exposure} recently highlighted the efficacy of MRT for reducing the effects of exposure bias. 
\section{Experimental setup}
\begin{table*}[ht]
\centering
\small
\begin{tabular}{l|l|cc|cc|}
   \cline{2-6}

 & &\textbf{de2en}  & \textbf{en2de}   & \textbf{es2en}  & \textbf{en2es}\\
   \cline{2-6}
   \footnotesize{1}& Baseline  & 38.8 &30.6 & 48.5  & 46.6 \\
  \footnotesize{2}&  MLE fine-tuning from 1 &40.9 & 32.5 & 48.5 & 46.0 \\

   \footnotesize{3}&  Checkpoint averaging  2 (en-de) / 1 (en-es) & 41.1 & 32.2& 48.5 &47.1  \\
   \cline{2-6}
   \footnotesize{4}&  MRT from 1  & 40.0 & 31.1 & \textbf{49.0} & 47.4 \\

   \footnotesize{5}&  MRT from 2 (en-de only)  &\textbf{41.3} & 32.9 & - &-\\
\footnotesize{6}& Checkpoint averaging 5 (en-de) / 4 (en-es) & \textbf{41.3} &\textbf{33.0} & 48.9 & \textbf{47.7}\\

   \cline{2-6}

\end{tabular}
\caption{Validation BLEU developing models used in English-German and English-Spanish  language pair submissions. Scores for single checkpoints unless indicated. MLE fine-tuning did not improve over the en-es baselines, so we do not use these models to initialise MRT.}\label{tab:ablation-results}
\end{table*}

\footnotetext[1]{\url{https://ufal.mff.cuni.cz/ufal\_medical\_corpus}} 
\footnotetext[2]{\citet{dusek2017khresmoi}}
\footnotetext[3]{\citet{neves2016scielo}}
\footnotetext[4]{\url{https://github.com/biomedical-translation-corpora/medline} \cite{yepes2017findings}}
\footnotetext[5]{\url{http://www.himl.eu/test-sets}}

\begin{table*}[ht]
\centering
\small
\begin{tabular}{|l|cc|cc|}
\hline
 &\textbf{de2en}  & \textbf{en2de}   & \textbf{es2en}  & \textbf{en2es}\\
 \hline
    MLE from baseline  & 41.1 & 32.2& - & -  \\

  MLE from baseline, no-title  &  41.4 &  31.8&-  &- \\

 MRT from: MLE (en-de) / baseline (en-es) & 41.3 &\textbf{33.0} & 48.9 & \textbf{47.7}\\
 MRT no-title from: MLE  no-title (en-de) / baseline (en-es) & \textbf{41.9} & 32.6 & \textbf{49.0} & 47.2\\
\hline
\end{tabular}
\caption{Validation BLEU developing models used in English-German and English-Spanish  language pair submissions. Scores for averaged checkpoints. MLE fine-tuning with either dataset did not improve over the en-es baselines.}\label{tab:notitles-results}
\end{table*}

\begin{table*}[ht]
\centering
\small
\begin{tabular}{|l|cccc|cccc|}
\hline
  & \multicolumn{2}{c}{\textbf{de2en}}  & \multicolumn{2}{c}{\textbf{en2de}}   & \multicolumn{2}{c}{\textbf{es2en}}  & \multicolumn{2}{c|}{\textbf{en2es}}\\
     &   \textbf{Dev} &  \textbf{Test} & \textbf{Dev}  &  \textbf{Test} &   \textbf{Dev} &  \textbf{Test} & \textbf{Dev}  &  \textbf{Test}\\ \hline
% following lines give scores to more places as given in official scores
%   MLE (all data)  & & 39.62 & & 32.88 & & 46.62 & & 45.72\\
%   MRT (no-titles data)  & & 39.63 & & 32.82 & & 46.40 & &46.72 \\
%   MRT (all data) & & 39.79 & & 33.18 & & 46.57 & & 46.62\\
   MLE (all data) (en-de) / Baseline (en-es)  & 41.1 & 39.6 &  32.2 & 32.9 & 48.5 & \textbf{46.6} & 47.1 & 45.7\\
   MRT (no-title data)  & \textbf{41.9} & 39.6 & 32.6 & 32.8 & \textbf{49.0} & 46.4 & 47.2&\textbf{46.7} \\
   MRT (all data) & 41.3 & \textbf{39.8} & \textbf{33.0} & \textbf{33.2} & 48.9 & \textbf{46.6} & 47.7& 46.6\\
   \hline
\end{tabular}
\caption{Validation and test BLEU for models used in English-German and English-Spanish  language pair submissions. Test results are for "OK sentences" as scored by the organizers.}\label{tab:submission-results}
\end{table*}
\label{ss:data}
\subsection{Data}
We report on two language pairs: English-Spanish (en-es) and English-German (en-de). Table \ref{tab:data} lists the data used to train our biomedical domain evaluation systems. For each language pair we use the same training data in both directions, and preprocess all data with Moses tokenization, punctuation normalization and truecasing.
 We use a 32K-merge joint source-target BPE vocabulary \cite{sennrich2016subword} learned on the pre-training data. 
 
All of our submitted approaches involve fine-tuning pre-trained models. We initialise fine-tuning with the strong biomedical domain models that formed our `run 1' submission for the WMT19 biomedical translation task. Details of data preparation and training for these models are discussed in \citet{saunders-etal-2019-ucam}.

We fine-tune these models on Medline abstracts data, validating on test sets from the 2019 Biomedical task. For these  we concatenate the src-trg and trg-src 2019 test sets for each language pair, and select only the `OK' aligned sentences as annotated by the organizers. 

Before fine-tuning we carry out detected language filtering on the Medline abstracts fine-tuning data using the Python LangDetect package\footnote{\url{https://pypi.org/project/langdetect/}}. We find LangDetect has a tendency to incorrectly label short sentences or those with rare vocabulary (very common in Medline) as a random language. For each language pair we therefore filter out only sentences where LangDetect identifies the source sentence as belonging to the target language, and vice versa. 

We then use a series of simple heuristics to further filter the parallel datasets, removing duplicate sentence pairs, those with source/target length ratio of $<$ 1:3.5 or $>$ 3.5:1, and sentences with $>$ 120 tokens. For the more aggressively-filtered `no-title' experiments we additionally remove all lines containing multiple tokens in square brackets, which in medical writing are used to denote the English translation of a non-English article's title \citep{patrias2007citing}. This leaves 27.3K sentence pairs for en-de and 64.8K for en-es: about 96\% of the filtered data in both cases.

\subsection{Model hyperparameters and training}
We use the Tensor2Tensor implementation of the Transformer model with the \texttt{transformer\_big} setup for all NMT models \cite{tensor2tensor}. We use the same effective batch size of 4k tokens for both MLE and doc-MRT. Because of model size constraints and the need to sample multiple targets for doc-MRT, we achieve the 4k effective batch size by accumulating gradients \citep{saunders2018multi} over every 4 batches of 1k tokens for MLE and every 16 batches of 256 tokens for doc-MRT.

For doc-MRT we use sampling temperature $\tau=0.3$, smoothing parameter $\alpha=0.6$ and $N=8$ samples per sentence, which gave the best results for our doc-MRT experiments in \citet{saunders-etal-2020-using}.

For each approach we fine-tune on a single GPU, saving checkpoints every 1K updates, until fine-tuning validation set BLEU fails to improve for 3 consecutive checkpoints. Generally this took about 5K updates. We then perform checkpoint averaging \cite{sys-amu-wmt16} over the final 3 checkpoints to obtain the final model.

\subsection{Inference}

For the 2020 submissions, we additionally split any test lines containing multiple sentences before inference using the Python NLTK package\footnote{\url{https://pypi.org/project/nltk/} sentence splitter}, translate the split sentences separately, then remerged. We found this gave noticeable improvements in quality for the few sentences it applied to. In all cases we decode with beam size 4 using SGNMT \cite{stahlberg2017sgnmt}. Test scores are as provided by the organizers for "OK" sentences using Moses tokenization and the multi-eval tool. Validation scores are for case-insensitive, detokenized text obtained using SacreBLEU\footnote{SacreBLEU signature: \texttt{BLEU+case.lc+numrefs.1\\+smooth.exp+tok.13a+version.1.2.11}} \citep{post2018call}.

% \begin{figure*}[ht]
% \centering
% \small
% \includegraphics[width=\linewidth]{transfer-learning.png}
% \caption{Transfer learning for es2en domains. Left: standard transfer learning improves performance from a smaller (health) to a larger (pre-trained biomedical) domain. Right: returning to the original domain after transfer learning provides further gains on health.}
% \label{fig:transfer}
% \end{figure*}

\subsection{Results}
We first assess the impact of small-domain adaptation to the full title-included Medline training set. Results in Table \ref{tab:ablation-results} show that small-domain MLE can lead to over-fitting and reduced performance (en-es) but also significant gains (en-de). Further fine-tuning with doc-MRT improved performance relative to the best MLE model for all translation directions by up to 0.8 BLEU when comparing with or without checkpoint averaging. While checkpoint averaging slightly decreased validation set performance for en2de MLE, we use it in all cases since it reduces sensitivity to randomness in training \citep{popel2018training}.  

In Table \ref{tab:notitles-results} we explore the impact of fine-tuning only on aggressively filtered `no-title' data. This does noticeably improve performance for de2en, with a very small improvement for es2en. Since the added information in `title' sentences is on the English side, this suggests that target training sentence quality impacts both MLE and MRT performance. However, removing these sentences entirely results in a noticeable performance decrease for the en2de and en2es models, demonstrating that they can be valuable training examples.

We submitted three runs to the WMT20 biomedical task for each language pair. For en-de run 1 was the baseline model fine-tuned on MLE with all data, while for en-es we submitted the checkpoint averaged baseline as MLE fine-tuning did not improve dev set performance. Run 2 was the run 1 model fine-tuned with doc-MRT on no-title data. Run 3 was the run 1 model fine-tuned with doc-MRT on all Medline abstract data. Table \ref{tab:submission-results} gives scores for these submitted models. 

Our best runs achieve the best and second-best results among all systems for en2es and es2en respectively as reported by the organizers. For en-de our test scores are further behind other systems, perhaps indicating that the baseline system could have been stronger before fine-grained adaptation. This is also indicated by the strong improvement of these models under simple MLE.

We submitted the MRT model on no-title data instead of the MLE on no-title data because MLE optimization did not improve over the baseline for en-es or en-es, with or without title lines, whereas MRT fine-tuning did. We also wanted to further examine whether MRT was robust enough to benefit from `noisy' data like the title lines, or whether cleaner no-title training data was more useful. In fact both forms of doc-MRT performed similarly on the test data, except in the case of en2de, where `no-title' MRT scored 0.4 BLEU worse -- further confirmation that source sentences with more information than the gold target can benefit MRT. We note that a MRT run was the best run or tied best run in all cases.

For the test runs, we additionally experimented with simply removing square bracket tokens from source sentences, since these could act as `triggering' tokens for title sentences. This did seem to improve translations for the sentences it applied to, but is clearly not applicable to all forms of exposure bias, since it requires knowledge of all behaviours that could trigger exposure bias. MRT does not require such knowledge, but still reduces the effects of exposure bias.

\section{Conclusions}
Our WMT20 Biomedical submission investigates improvements on the English-German and English-Spanish language pairs under a single strong model. In particular, we focus on the behaviour of models trained on sentences with some predictable irregularities. We find that aggressively filtering target sentences can help overall performance, but that aggressively filtering source sentence tends to hurt performance. We also find that Minimum Risk Training can benefit from imperfectly aligned training examples while reducing the effects of exposure bias.

\section*{Acknowledgments}
This work was supported by EPSRC grants EP/M508007/1 and EP/N509620/1 and has been performed using resources provided by the Cambridge Tier-2 system operated by the University of Cambridge Research Computing Service\footnote{\url{http://www.hpc.cam.ac.uk}} funded by EPSRC Tier-2 capital grant EP/P020259/1.

\bibliographystyle{acl_natbib}
\bibliography{refs}

\begin{thebibliography}{28}
\expandafter\ifx\csname natexlab\endcsname\relax\def\natexlab#1{#1}\fi

\bibitem[{Bawden et~al.(2019)Bawden, Bretonnel~Cohen, Grozea, Jimeno~Yepes,
  Kittner, Krallinger, Mah, Neveol, Neves, Soares, Siu, Verspoor, and
  Vicente~Navarro}]{bawden-etal-2019-findings}
Rachel Bawden, Kevin Bretonnel~Cohen, Cristian Grozea, Antonio Jimeno~Yepes,
  Madeleine Kittner, Martin Krallinger, Nancy Mah, Aurelie Neveol, Mariana
  Neves, Felipe Soares, Amy Siu, Karin Verspoor, and Maika Vicente~Navarro.
  2019.
\newblock \href {https://doi.org/10.18653/v1/W19-5403} {Findings of the {WMT}
  2019 biomedical translation shared task: Evaluation for {MEDLINE} abstracts
  and biomedical terminologies}.
\newblock In \emph{Proceedings of the Fourth Conference on Machine Translation
  (Volume 3: Shared Task Papers, Day 2)}, pages 29--53, Florence, Italy.
  Association for Computational Linguistics.

\bibitem[{Bengio et~al.(2015)Bengio, Vinyals, Jaitly, and
  Shazeer}]{bengio2015scheduled}
Samy Bengio, Oriol Vinyals, Navdeep Jaitly, and Noam Shazeer. 2015.
\newblock Scheduled sampling for sequence prediction with recurrent neural
  networks.
\newblock In \emph{Advances in Neural Information Processing Systems}, pages
  1171--1179.

\bibitem[{Chu et~al.(2017)Chu, Dabre, and Kurohashi}]{chu2017empirical}
Chenhui Chu, Raj Dabre, and Sadao Kurohashi. 2017.
\newblock An empirical comparison of domain adaptation methods for neural
  machine translation.
\newblock In \emph{Proceedings of the 55th Annual Meeting of the Association
  for Computational Linguistics (Volume 2: Short Papers)}, pages 385--391.

\bibitem[{Du{\v s}ek et~al.(2017)Du{\v s}ek, Haji{\v c}, Hlav{\'a}{\v
  c}ov{\'a}, Libovick{\'y}, Pecina, Tamchyna, and Ure{\v
  s}ov{\'a}}]{dusek2017khresmoi}
Ond{\v r}ej Du{\v s}ek, Jan Haji{\v c}, Jaroslava Hlav{\'a}{\v c}ov{\'a},
  Jind{\v r}ich Libovick{\'y}, Pavel Pecina, Ale{\v s} Tamchyna, and Zde{\v
  n}ka Ure{\v s}ov{\'a}. 2017.
\newblock \href {http://hdl.handle.net/11234/1-2122} {Khresmoi summary
  translation test data 2.0}.
\newblock {LINDAT}/{CLARIN} digital library at the Institute of Formal and
  Applied Linguistics ({{\'U}FAL}), Faculty of Mathematics and Physics, Charles
  University.

\bibitem[{Edunov et~al.(2018)Edunov, Ott, Auli, Grangier
  et~al.}]{edunov2018classical}
Sergey Edunov, Myle Ott, Michael Auli, David Grangier, et~al. 2018.
\newblock Classical structured prediction losses for sequence to sequence
  learning.
\newblock In \emph{Proceedings of the 2018 Conference of the North American
  Chapter of the Association for Computational Linguistics: Human Language
  Technologies, Volume 1 (Long Papers)}, volume~1, pages 355--364.

\bibitem[{Farajian et~al.(2017)Farajian, Turchi, Negri, and
  Federico}]{farajian-etal-2017-multi}
M.~Amin Farajian, Marco Turchi, Matteo Negri, and Marcello Federico. 2017.
\newblock \href {https://doi.org/10.18653/v1/W17-4713} {Multi-domain neural
  machine translation through unsupervised adaptation}.
\newblock In \emph{Proceedings of the Second Conference on Machine
  Translation}, pages 127--137, Copenhagen, Denmark. Association for
  Computational Linguistics.

\bibitem[{Junczys-Dowmunt et~al.(2016)Junczys-Dowmunt, Dwojak, and
  Sennrich}]{sys-amu-wmt16}
Marcin Junczys-Dowmunt, Tomasz Dwojak, and Rico Sennrich. 2016.
\newblock \href {http://www.aclweb.org/anthology/W16-2316} {The {AMU-UEDIN}
  submission to the {WMT16} news translation task: Attention-based {NMT} models
  as feature functions in phrase-based {SMT}}.
\newblock In \emph{Proceedings of the First Conference on Machine Translation},
  pages 319--325, Berlin, Germany. Association for Computational Linguistics.

\bibitem[{Kothur et~al.(2018)Kothur, Knowles, and
  Koehn}]{kothur-etal-2018-document}
Sachith Sri~Ram Kothur, Rebecca Knowles, and Philipp Koehn. 2018.
\newblock \href {https://doi.org/10.18653/v1/W18-2708} {Document-level
  adaptation for neural machine translation}.
\newblock In \emph{Proceedings of the 2nd Workshop on Neural Machine
  Translation and Generation}, pages 64--73, Melbourne, Australia. Association
  for Computational Linguistics.

\bibitem[{Li et~al.(2018)Li, Zhang, and Zong}]{li-etal-2018-one}
Xiaoqing Li, Jiajun Zhang, and Chengqing Zong. 2018.
\newblock \href {https://www.aclweb.org/anthology/L18-1146} {One sentence one
  model for neural machine translation}.
\newblock In \emph{Proceedings of the Eleventh International Conference on
  Language Resources and Evaluation ({LREC} 2018)}, Miyazaki, Japan. European
  Language Resources Association (ELRA).

\bibitem[{Luong and Manning(2015)}]{luong2015stanford}
Minh-Thang Luong and Christopher~D Manning. 2015.
\newblock {Stanford Neural Machine Translation systems for spoken language
  domains}.
\newblock In \emph{Proceedings of the International Workshop on Spoken Language
  Translation}, pages 76--79.

\bibitem[{Michel and Neubig(2018)}]{michel-neubig-2018-extreme}
Paul Michel and Graham Neubig. 2018.
\newblock \href {https://doi.org/10.18653/v1/P18-2050} {Extreme adaptation for
  personalized neural machine translation}.
\newblock In \emph{Proceedings of the 56th Annual Meeting of the Association
  for Computational Linguistics (Volume 2: Short Papers)}, pages 312--318,
  Melbourne, Australia. Association for Computational Linguistics.

\bibitem[{Neubig(2016)}]{neubig-2016-lexicons}
Graham Neubig. 2016.
\newblock \href {https://www.aclweb.org/anthology/W16-4610} {Lexicons and
  minimum risk training for neural machine translation: {NAIST}-{CMU} at
  {WAT}2016}.
\newblock In \emph{Proceedings of the 3rd Workshop on {A}sian Translation
  ({WAT}2016)}, pages 119--125, Osaka, Japan. The COLING 2016 Organizing
  Committee.

\bibitem[{Neves et~al.(2016)Neves, Jimeno-Yepes, and
  N{\'e}v{\'e}ol}]{neves2016scielo}
Mariana~L Neves, Antonio Jimeno-Yepes, and Aur{\'e}lie N{\'e}v{\'e}ol. 2016.
\newblock {The ScieLO Corpus: a Parallel Corpus of Scientific Publications for
  Biomedicine}.
\newblock In \emph{LREC}.

\bibitem[{Patrias and Wendling(2007)}]{patrias2007citing}
Karen Patrias and Dan Wendling. 2007.
\newblock Citing medicine: the nlm style guide for authors, editors.
\newblock \emph{and publishers. Bethesda, MD: National Library of Medicine.
  Retrieved June}, 27:2011.

\bibitem[{Popel and Bojar(2018)}]{popel2018training}
Martin Popel and Ond{\v{r}}ej Bojar. 2018.
\newblock Training tips for the transformer model.
\newblock \emph{The Prague Bulletin of Mathematical Linguistics},
  110(1):43--70.

\bibitem[{Post(2018)}]{post2018call}
Matt Post. 2018.
\newblock {A call for clarity in reporting BLEU scores}.
\newblock \emph{CoRR}, abs/1804.08771.

\bibitem[{Ranzato et~al.(2016)Ranzato, Chopra, Auli, and
  Zaremba}]{ranzanto16sequencelevel}
Marc'Aurelio Ranzato, Sumit Chopra, Michael Auli, and Wojciech Zaremba. 2016.
\newblock Sequence level training with recurrent neural networks.
\newblock In \emph{ICLR}.

\bibitem[{Saunders and Byrne(2020)}]{saunders-byrne-2020-reducing}
Danielle Saunders and Bill Byrne. 2020.
\newblock \href {https://www.aclweb.org/anthology/2020.acl-main.690} {Reducing
  gender bias in neural machine translation as a domain adaptation problem}.
\newblock In \emph{Proceedings of the 58th Annual Meeting of the Association
  for Computational Linguistics}, pages 7724--7736, Online. Association for
  Computational Linguistics.

\bibitem[{Saunders et~al.(2019)Saunders, Stahlberg, and
  Byrne}]{saunders-etal-2019-ucam}
Danielle Saunders, Felix Stahlberg, and Bill Byrne. 2019.
\newblock \href {https://doi.org/10.18653/v1/W19-5421} {{UCAM} biomedical
  translation at {WMT}19: Transfer learning multi-domain ensembles}.
\newblock In \emph{Proceedings of the Fourth Conference on Machine Translation
  (Volume 3: Shared Task Papers, Day 2)}, pages 169--174, Florence, Italy.
  Association for Computational Linguistics.

\bibitem[{Saunders et~al.(2020)Saunders, Stahlberg, and
  Byrne}]{saunders-etal-2020-using}
Danielle Saunders, Felix Stahlberg, and Bill Byrne. 2020.
\newblock \href {https://www.aclweb.org/anthology/2020.acl-main.693} {Using
  context in neural machine translation training objectives}.
\newblock In \emph{Proceedings of the 58th Annual Meeting of the Association
  for Computational Linguistics}, pages 7764--7770, Online. Association for
  Computational Linguistics.

\bibitem[{Saunders et~al.(2018)Saunders, Stahlberg, de~Gispert, and
  Byrne}]{saunders2018multi}
Danielle Saunders, Felix Stahlberg, Adri{\`a} de~Gispert, and Bill Byrne. 2018.
\newblock Multi-representation ensembles and delayed sgd updates improve
  syntax-based nmt.
\newblock In \emph{Proceedings of the 56th Annual Meeting of the Association
  for Computational Linguistics (Volume 2: Short Papers)}, pages 319--325.

\bibitem[{Sennrich et~al.(2016)Sennrich, Haddow, and
  Birch}]{sennrich2016subword}
Rico Sennrich, Barry Haddow, and Alexandra Birch. 2016.
\newblock \href {https://doi.org/10.18653/v1/P16-1162} {{Neural Machine
  Translation of Rare Words with Subword Units}}.
\newblock In \emph{Proceedings of the 54th Annual Meeting of the Association
  for Computational Linguistics}, volume~1, pages 1715--1725.

\bibitem[{Shen et~al.(2016)Shen, Cheng, He, He, Wu, Sun, and
  Liu}]{shen2016minimum}
Shiqi Shen, Yong Cheng, Zhongjun He, Wei He, Hua Wu, Maosong Sun, and Yang Liu.
  2016.
\newblock {Minimum Risk Training for Neural Machine Translation}.
\newblock In \emph{Proceedings of the 54th Annual Meeting of the Association
  for Computational Linguistics}, volume~1, pages 1683--1692.

\bibitem[{Stahlberg et~al.(2017)Stahlberg, Hasler, Saunders, and
  Byrne}]{stahlberg2017sgnmt}
Felix Stahlberg, Eva Hasler, Danielle Saunders, and Bill Byrne. 2017.
\newblock \href {https://doi.org/10.18653/v1/D17-2005} {{SGNMT--A Flexible NMT
  Decoding Platform for Quick Prototyping of New Models and Search
  Strategies}}.
\newblock In \emph{Proceedings of the 2017 Conference on Empirical Methods in
  Natural Language Processing: System Demonstrations}, pages 25--30.

\bibitem[{Vaswani et~al.(2018)Vaswani, Bengio, Brevdo, Chollet, Gomez, Gouws,
  Jones, Kaiser, Kalchbrenner, Parmar, Sepassi, Shazeer, and
  Uszkoreit}]{tensor2tensor}
Ashish Vaswani, Samy Bengio, Eugene Brevdo, Francois Chollet, Aidan~N. Gomez,
  Stephan Gouws, Llion Jones, \L{}ukasz Kaiser, Nal Kalchbrenner, Niki Parmar,
  Ryan Sepassi, Noam Shazeer, and Jakob Uszkoreit. 2018.
\newblock {Tensor2Tensor for Neural Machine Translation}.
\newblock \emph{CoRR}, abs/1803.07416.

\bibitem[{Wang and Sennrich(2020)}]{wang-sennrich-2020-exposure}
Chaojun Wang and Rico Sennrich. 2020.
\newblock \href {https://www.aclweb.org/anthology/2020.acl-main.326} {On
  exposure bias, hallucination and domain shift in neural machine translation}.
\newblock In \emph{Proceedings of the 58th Annual Meeting of the Association
  for Computational Linguistics}, pages 3544--3552, Online. Association for
  Computational Linguistics.

\bibitem[{Xu et~al.(2019)Xu, Crego, and Senellart}]{xu2019lexical}
Jitao Xu, Josep Crego, and Jean Senellart. 2019.
\newblock Lexical micro-adaptation for neural machine translation.
\newblock In \emph{International Workshop on Spoken Language Translation}.

\bibitem[{Yepes et~al.(2017)Yepes, N{\'e}v{\'e}ol, Neves, Verspoor, Bojar,
  Boyer, Grozea, Haddow, Kittner, Lichtblau et~al.}]{yepes2017findings}
Antonio~Jimeno Yepes, Aur{\'e}lie N{\'e}v{\'e}ol, Mariana Neves, Karin
  Verspoor, Ondrej Bojar, Arthur Boyer, Cristian Grozea, Barry Haddow,
  Madeleine Kittner, Yvonne Lichtblau, et~al. 2017.
\newblock Findings of the wmt 2017 biomedical translation shared task.
\newblock In \emph{Proceedings of the Second Conference on Machine
  Translation}, pages 234--247.

\end{thebibliography}

\end{document}